\documentclass[11pt,a4paper]{article}
\usepackage[hyperref]{eacl2021}
\usepackage{times}
\usepackage{latexsym}

\usepackage{microtype}
\usepackage{graphicx}
\usepackage{multirow}
\usepackage{multicol}
\usepackage{mathtools}
\usepackage{amsmath}

\aclfinalcopy 
\def\aclpaperid{23} 

\title{User Factor Adaptation for User Embedding via Multitask Learning}

\author{Xiaolei Huang \\
  Computer Science \\
  University of Memphis \\
  \texttt{xiaolei.huang@memphis.edu} \\\And
  Michael J. Paul, Robin Burke \\
  Information Science \\
  University of Colorado Boulder \\
  \texttt{~~~~{mpaul,robin.burke}@colorado.edu} \\\AND
  Franck Dernoncourt \\
  Adobe Research \\
  \texttt{dernonco@adobe.com} \\\And
  Mark Dredze \\
  Computer Science \\
  Johns Hopkins University \\
  \texttt{mdredze@cs.jhu.edu} \\
  }

\date{}

\begin{document}
\maketitle

\begin{abstract}

Language varies across users and their interested fields in social media data:
words authored by a user across his/her interests may have different meanings (e.g., cool) or sentiments (e.g., fast).
However, most of the existing methods to train user embeddings ignore the variations across user interests, such as product and movie categories (e.g., drama vs. action).
In this study, we treat the user interest as domains and empirically examine how the user language can vary across the user factor in three English social media datasets.
We then propose a user embedding model to account for the language variability of user interests via a multitask learning framework.
The model learns user language and its variations without human supervision.
While existing work mainly evaluated the user embedding by extrinsic tasks, we propose an intrinsic evaluation via clustering and evaluate user embeddings by an extrinsic task, text classification.
The experiments on the three English-language social media datasets show that our proposed approach can generally outperform baselines via adapting the user factor.

\end{abstract}

\section{Introduction}

Language varies across user factors including user interests, demographic attributes, personalities, and latent factors from user history.
Research shows that language usage diversifies according to online user groups~\cite{volkova2013exploring}, which women were more likely to use the word \textit{weakness} in a positive way while men were the opposite.
In social media, the user interests can include topics of user reviews (e.g., home vs. health services in Yelp) and categories of reviewed items (electronic vs kitchen products in Amazon).
The ways that users express themselves depend on current contexts of user interests~\cite{oba2019modeling} that users may use the same words for opposite meanings and different words for the same meaning.
For example, online users can use the word ``fast'' to criticize battery quality of the electronic domain or praise medicine effectiveness of the medical products; users can also use the words ``cool'' to describe a property of AC products or express sentiments.

User embedding, which is to learn a fixed-length representation based on multiple user reviews of each user, can infer the user latent information into a unified vector space~\cite{benton2018learning, pan2019social}.
The inferred latent representations from online content can predict user profile~\cite{volkova2015inferring, wang2018cross, farnadi2018user, lynn2020hierarchical} and behaviors~\cite{zhang2015using, amir2017quantifying, benton2017multitask, ding2017multi}.
User embeddings can personalize classification models, and further improve model performance~\cite{tang2015learning, chen2016neural, yang2017overcoming, wu2018improving, zeng2019joint, huang2019deep}.
The representations of user language can help models better understand documents as global contexts.

However, existing user embedding methods \cite{amir2016modelling, benton2016learning, xing2017incorporating, pan2019social} mainly focus on extracting features from language itself while ignoring user interests.
Recent research has demonstrated that adapting the user factors can further improve user geolocation prediction \cite{miura2017unifying}, demographic attribute prediction \cite{farnadi2018user}, and sentiment analysis \cite{yang2017overcoming}.
\newcite{lynn2017human, huang2019neuraluser} treated the language variations as a domain adaptation problem and referred to this idea as \textit{user factor adaptation}.

In this study, we treat the user interest as \textit{domains} (e.g., restaurants vs. home services domains) and propose a multitask framework to model language variations and incorporate the user factor into user embeddings.
We focus on three online review datasets from Amazon, IMDb, and Yelp containing diverse behaviors conditioned on user interests, which refer to genres of reviewed items.
For example, if any Yelp users have reviews on items of the home services, then their user interests will include the home services.

We start with exploring how the user factor, user interest, can cause language and classification variations in Section~\ref{chap4:subsec:analysis}.
We then propose our user embedding model that adapts the user interests using a multitask learning framework in Section~\ref{chap4:subsec:model2}.
Research~\cite{pan2019social} generally evaluates the user embedding via downstream tasks, but user annotations sometimes are hard to obtain and those evaluations are extrinsic instead of intrinsic tasks.
For example, the MyPersonality~\cite{kosinski2015facebook} that was used in previous work~\cite{ding2017multi, farnadi2018user, pan2019social} is no longer available, and an extrinsic task is to evaluate if user embeddings can help text classifiers.
Research~\cite{schnabel2015evaluation} suggests that the intrinsic evaluation including clustering is better than the extrinsic evaluation for controlling less hyperparameters.
We propose an intrinsic evaluation for user embedding, which can provide a new perspective for testing future experiments.
We show that our user-factor-adapted user embedding can generally outperform the existing methods on both intrinsic and extrinsic tasks.

\section{Data}
\label{chap4:subsec:data2}

We collected English reviews of Amazon (health product), IMDb and Yelp from the publicly available sources~\cite{he2016ups, yelp_2019, imdb2020dataset}.
For the IMDb dataset, we included English movies produced in the US from 1960 to 2019.
Each review associates with its author and the \textit{rated item}, which refers to a movie in the IMDb data, a business unit in the Yelp data and a product in the Amazon data.
To keep consistency in each dataset, we retain top 4 frequent genres of rated items and the review documents with no less than 10 tokens.\footnote{The top 4 rated categories of Amazon-Health, IMDb and Yelp are [sports nutrition, sexual wellness, shaving \& hair removal, vitamins \& dietary supplements], [comedy, thriller, drama, action] and [restaurants, health \& medical, home services, beauty \& spas] respectively.}
We dropped non-English review documents by the language detector~\cite{lui2012langid}, lowercased all tokens and tokenized the corpora using NLTK~\cite{bird2004nltk}.
The review datasets have different score scales. We normalize the scales and encode each review score into three discrete categories: positive ($>3$ for the Yelp and Amazon, $>6$ for the IMDb), negative ($<3$ for the Yelp and Amazon, $<5$ for the IMDb) and neutral.
Table~\ref{chap4:tab:data2} shows a summary of the datasets.

\subsection{Privacy Considerations}
To protect user privacy, we anonymize all user-related information via hashing, and our experiments only use publicly available datasets for research demonstration.
Any URLs, hashtags and capitalized English names were removed.
Due to the potential sensitivity of user reviews, we only use information necessary for this study.
We do not use any user profile in our experiments, except, our evaluations use anonymized author ID of each review entry for training user embeddings.
We will not release any private user reviews associated with user identities.
Instead, we will open source our source code and provide instructions of how to access the public datasets in enough detail so that our proposed method can be replicated.

\begin{table*}[htp]
\centering
    \begin{tabular}{c||cccc||ccc}
    Data & Users & Docs & Rated Items & Tokens & Train & Dev & Test \\\hline\hline
    Amazon-Health & 11,438 & 80,592 & 3,822 & 127 & 64,474 & 8,060 & 8,061 \\
    IMDb & 6,089 & 123,184 & 642 & 187 & 98,548 & 12,319 & 12,320 \\
    Yelp & 76,323 & 551,695 & 9,327 & 152 & 441,357 & 55,170 & 55,171 \\
    \end{tabular}
\caption{Statistical summary of the Amazon, Yelp and IMDb review datasets. Amazon-Health refers to health-related reviews. Tokens mean the number of average tokens per document. We present the data split for the evaluation task of text classification on the right side.}
\label{chap4:tab:data2}
\end{table*}

\section{Exploratory Analysis of User Variations}
\label{chap4:subsec:analysis}

Language varies across user factors such as user interests \cite{oba2019modeling}, demographic attributes \cite{huang2019neuraluser}, social relations \cite{yang2017overcoming, lin2020jnet}.
In this section, our goal is to quantitatively analyze whether the user interests cause user language variations, which can reduce effectiveness and robustness of user embeddings.
We approach this by two analysis tasks, first by measuring word feature similarity based on user interests, and second by examining how classifier performance depends on the grouped user interests in which the model is trained and applied.

\subsection{Word Usage Variations}

\begin{figure*}[htp]
    \centering
    \includegraphics[width=0.3\textwidth]{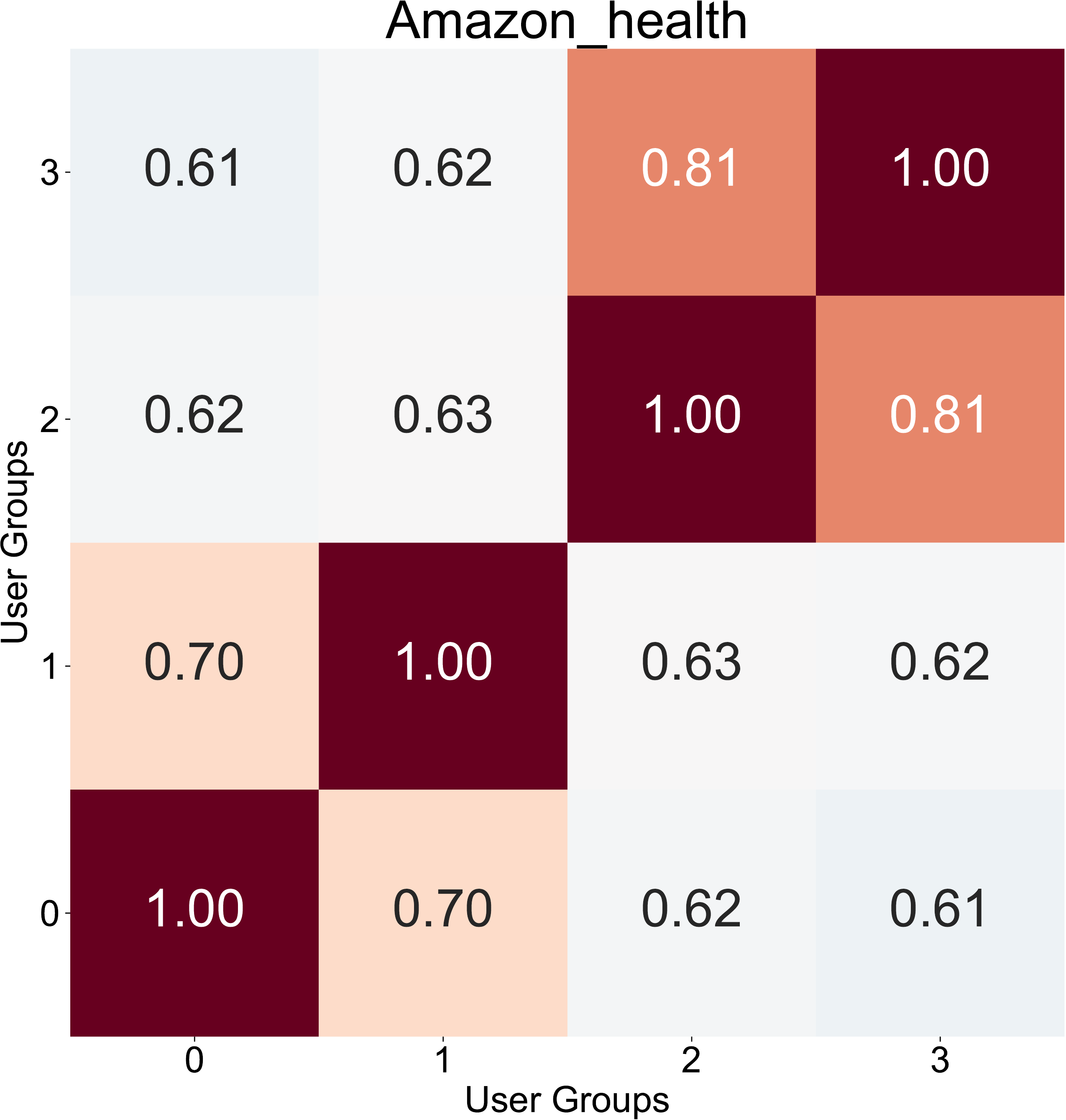}
    \includegraphics[width=0.3\textwidth]{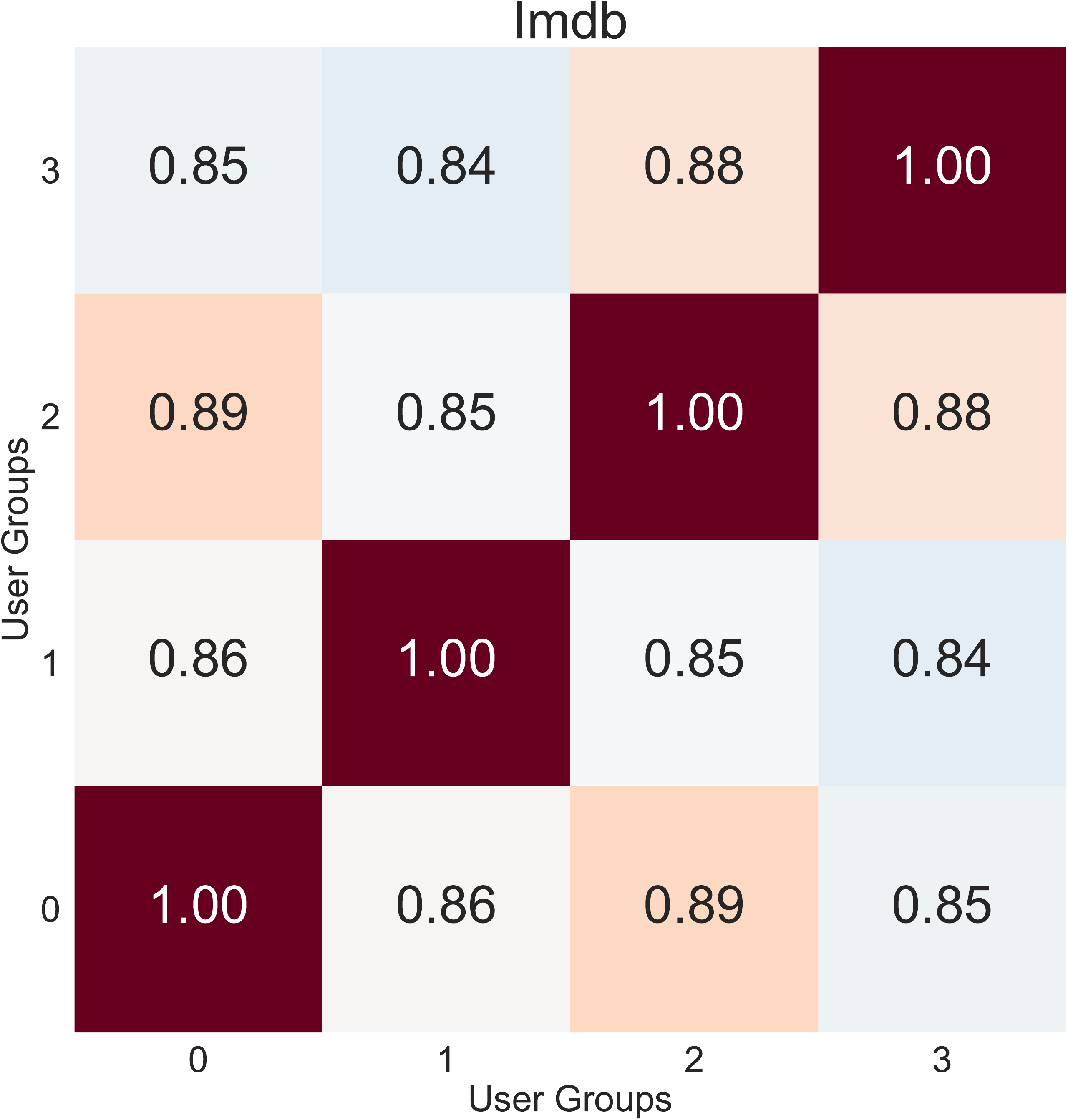}
    \includegraphics[width=0.3\textwidth]{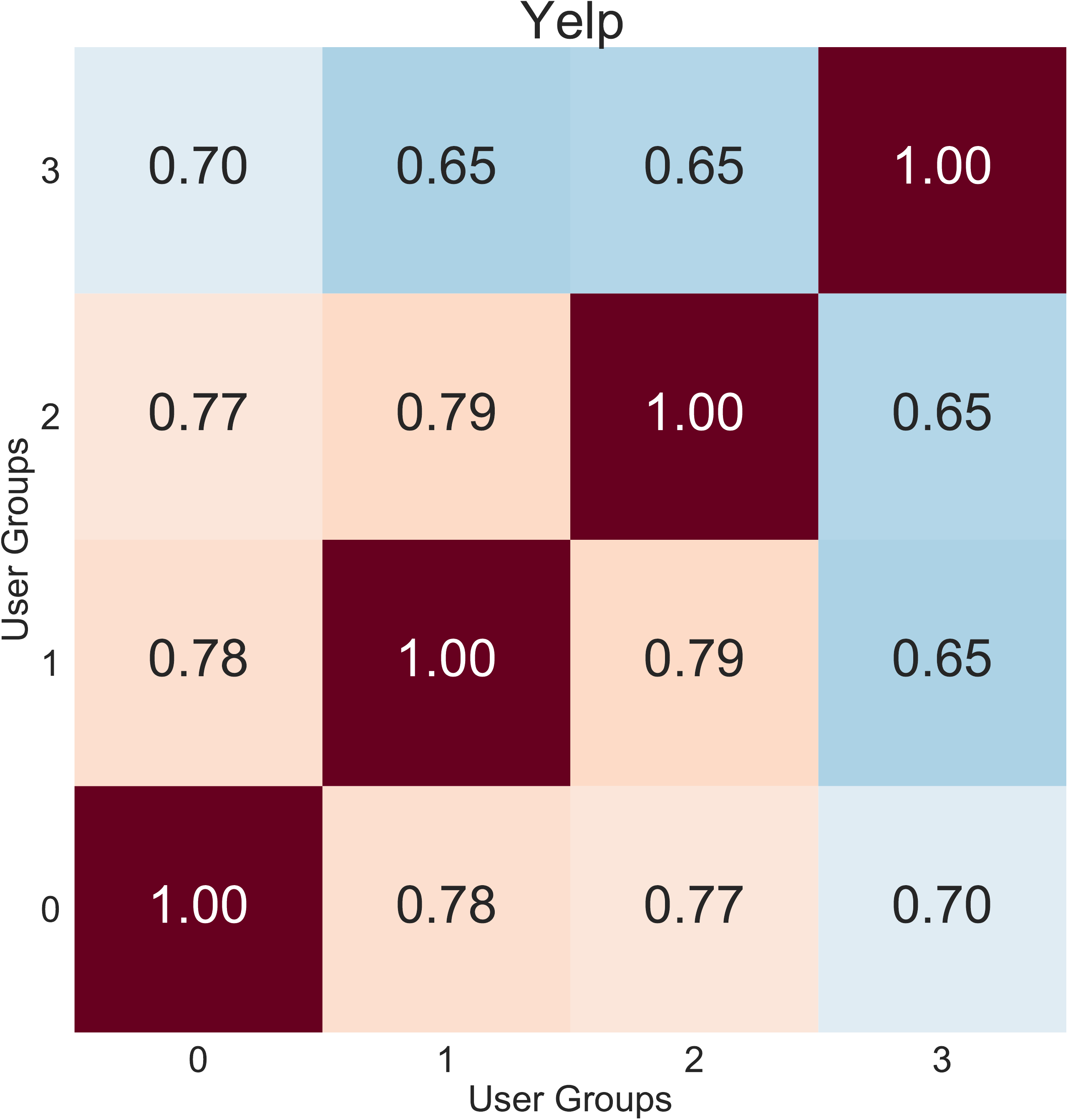}
    \caption{Word feature overlaps between every two user groups. A value of 1 means no variations of top features between two user groups, while values less than 1 indicate more feature variations.}
    \label{chap4:fig:word_user}
\end{figure*}

Existing methods mainly infer user embeddings from features of text contents~\cite{pan2019social}.
Therefore, word usage variations across user interests will change word distributions and further impact the stability of user embeddings.
We aim to test whether there are language variations across the user interests in our datasets and how strong they are.

We consider the word usage as it relates to user embeddings by estimating the overlap of top word features across the genres of rated items, the categories of reviewed products in Amazon, business units in Yelp and movies in IMDb.
To solve data sparsity caused by single user preference, we grouped users and therefore their generated documents according to genres of user reviewed items.
We refer to this as genre domains.
We build a unified feature vectorizer~\cite{pedregosa2011scikit} with TF-IDF weighted $n$-gram features ($n \in \{1,2,3\}$), removing features that appeared in less than 2 documents.
We rank and select the top 1000 word features for each genre domain by mutual information.
We then compute the intersection percentage between every two genre domains: let $F_0$ is the set of top features for one genre domain and $F_1$ is the set of top features for the other domain, then the overlap is $|F_0 \cap F_1|/1000$.

We show the results in Figure~\ref{chap4:fig:word_user}.
The overlap varies significantly across genre domains.
This indicates that the word usage and its contexts of users change across user interests and preferences. 
Since the training of user embeddings relies heavily on the language features of users, this suggests that it is important to consider the language variations in user interests for the user embeddings.

\subsection{Classification Performance Variations}

\begin{figure*}[htp]
\centering
\includegraphics[width=0.3\textwidth]{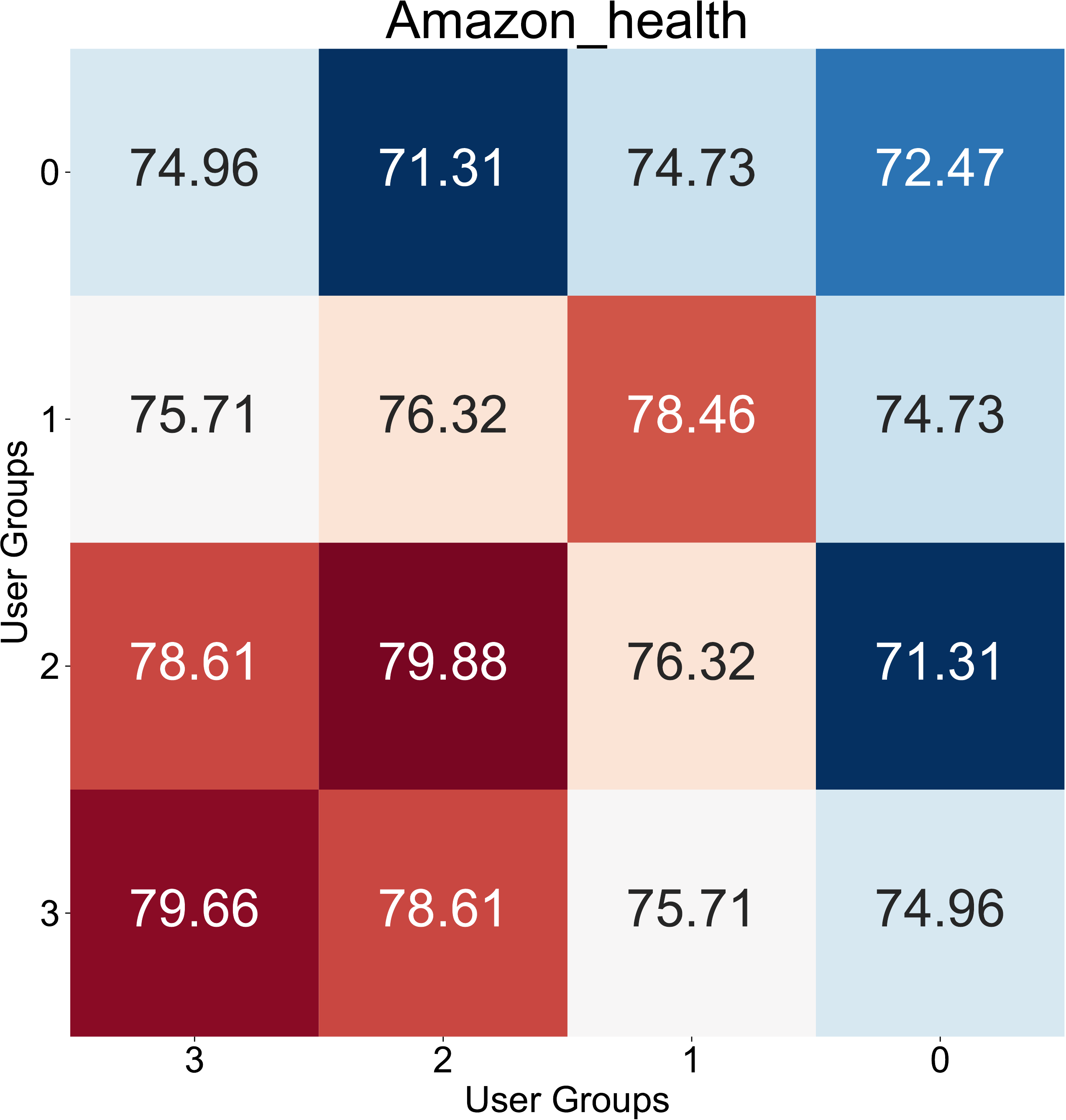}
\includegraphics[width=0.3\textwidth]{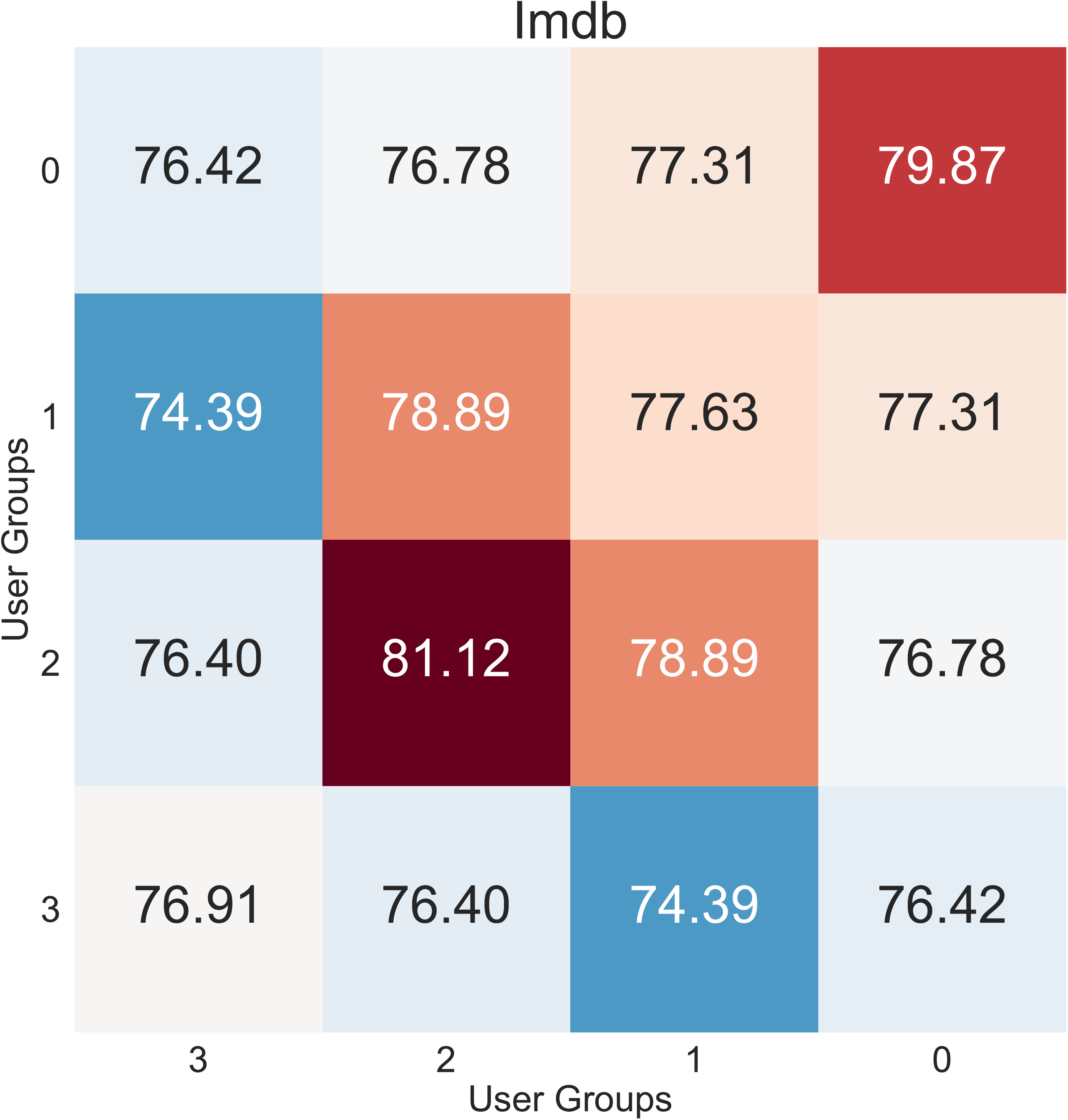}
\includegraphics[width=0.3\textwidth]{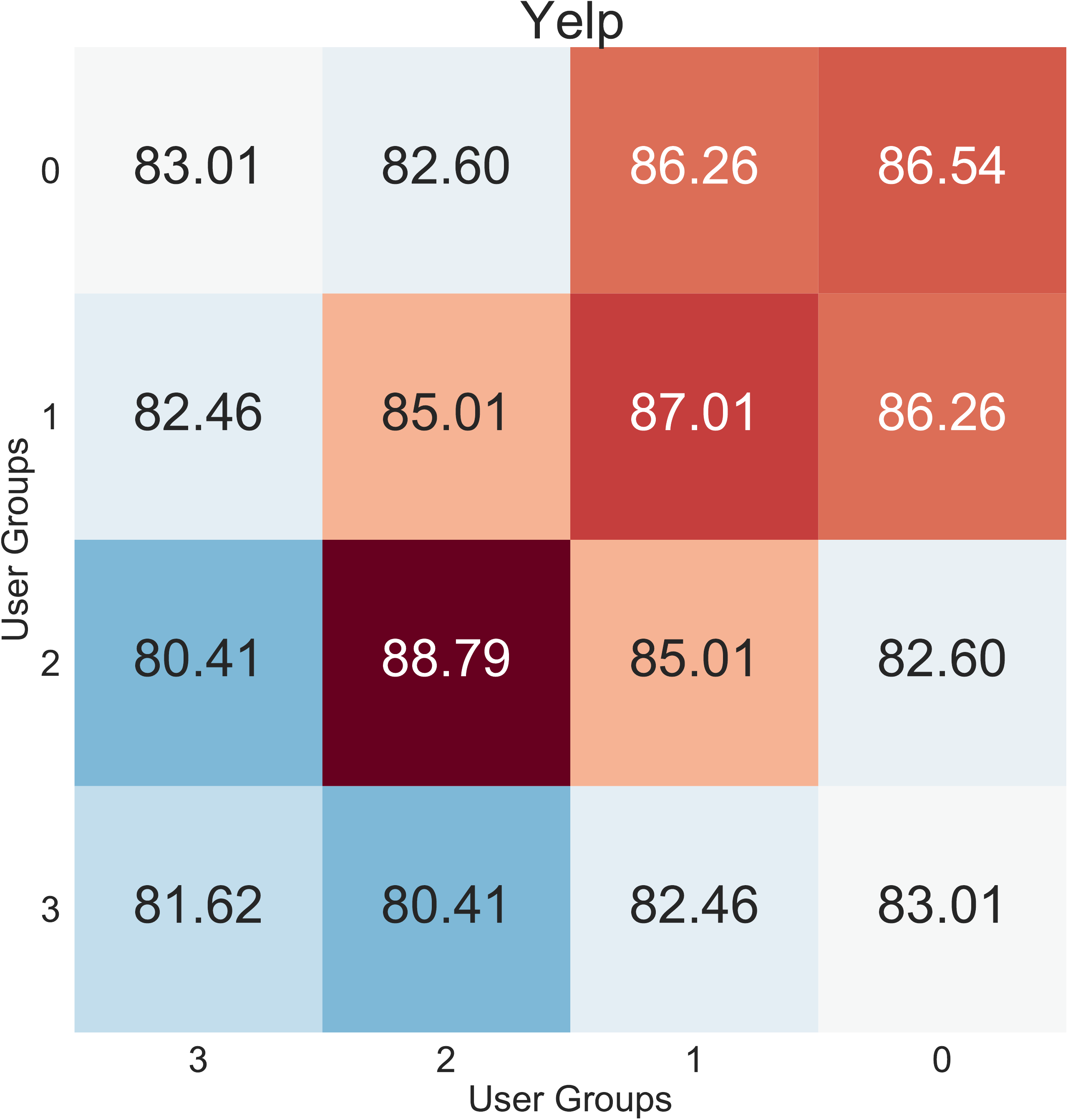}
\caption{Document classification performance when training and testing on different groups of users. The datasets come from Amazon health, IMDb and Yelp reviews. Darker red indicates better classification performance, while darker blue means worse performance.}
\label{chap4:fig:group}
\end{figure*}

User embeddings are effective to understand user behaviors in the classification setting~\cite{amir2016modelling, ding2018predicting}.
Research has found that combining user and document representations can benefit classification performance~\cite{chen2016learning, li2018document, yuan2019neural}.
We explore how the language variations in user interests can affect classification models.

We conduct an analysis by training and testing classifiers that group users by the categories of reviewed items.
We first group items and users according to item genres, which can be treated as different domains of user interests. 
For each domain, we downsampled documents, users and items within each group to match their numbers in the smallest group, so that classification performance differences are not due to data sizes of document, user and item.
For each grouped documents, we shuffle and split the data into training (80\%) and test (20\%) sets. We train logistic regression classifiers with default hyperparameters from scikit-learn~\cite{pedregosa2011scikit} using TF-IDF weighted uni-, bi- and tri-gram features. 
We report weighted F1 scores across grouped users and show the results in Figure~\ref{chap4:fig:group}.

We can observe that classification performance varies across the grouped users.
Higher performance variations between in- and out- user groups suggest higher user variations and vice versa.
If no variations of user language exist, the performance of classifiers should be similar across the domains.
The performance variations suggest that user behaviors vary across the categories of user interests. 
We can also observe that classification models generally perform better when tests within the same user groups while worse in the other user groups.
This suggests a variability connection between the user interests and language usage, which derives user embeddings.



\section{Multitask User Embedding}
\label{chap4:subsec:model2}

\begin{figure*}[t]
\centering
\includegraphics[width=.9\textwidth]{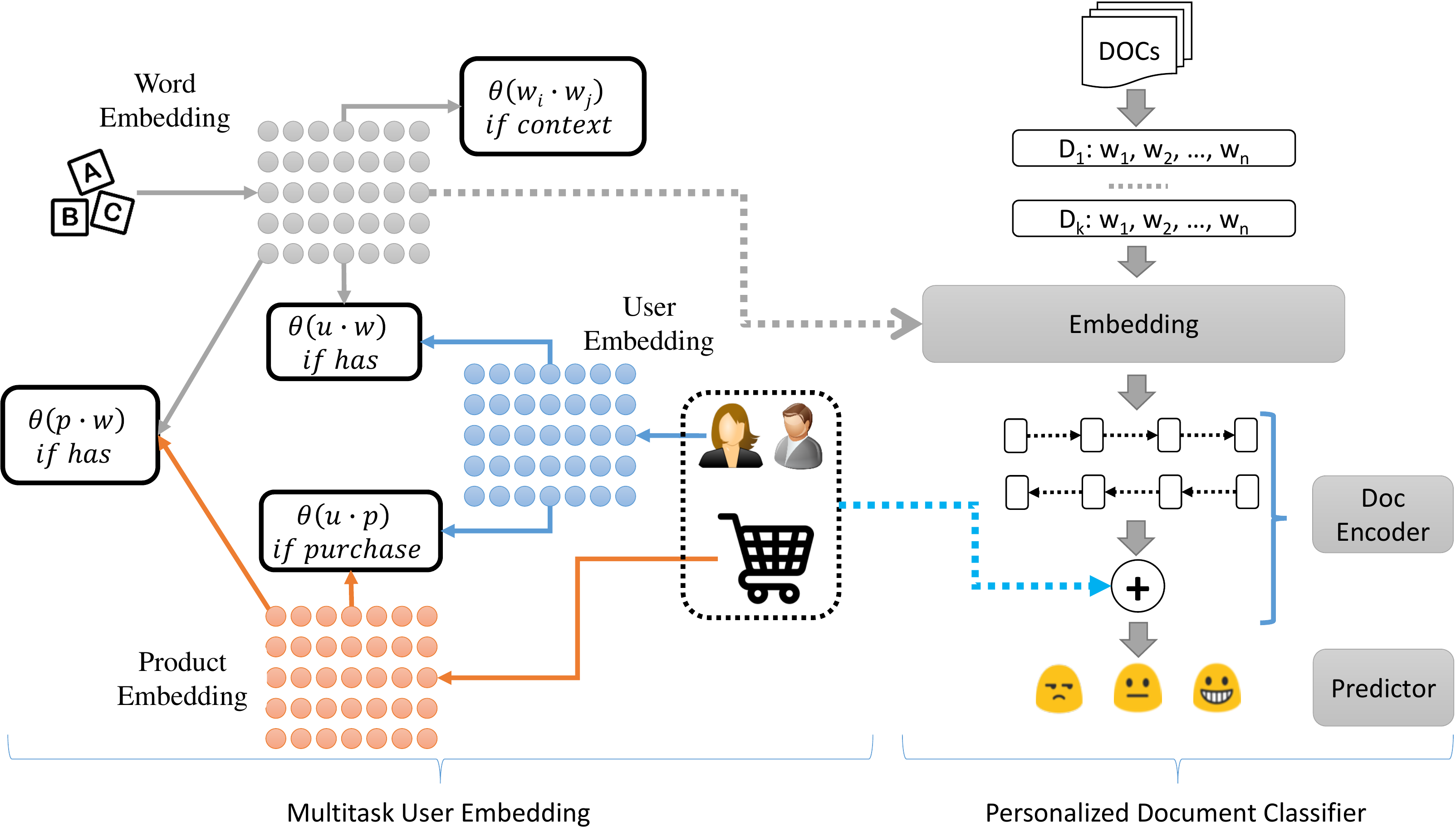}
\caption{Illustrations of User Embedding via multitask learning framework on the left and personalized document classifiers using trained embedding models on the right. The arrows and their colors refer to the input directions and input sources respectively. We use the logos of people, shopping cart and ABC to represent users, reviewed items and word inputs. The $\bigoplus$ is the concatenation operation.}
\label{chap4:fig:uemb_diagram}
\end{figure*}

We present the architecture of our proposed model in Figure~\ref{chap4:fig:uemb_diagram} on the left.
Methods~\cite{pan2019social} to train text-based user embedding mainly focus on the user-generated documents while ignoring user factors, the user interests. 
A close work to ours only trained user embeddings by predicting if users co-occurred with sampled words~\cite{amir2017quantifying}.
We extend this line of work by adapting user interests into modeling steps.
The proposed unsupervised model trains four joint tasks based on the Skip-Gram~\cite{mikolov2013distributed}: word and word, user and word, item and word, and user and item. 
Note that we do not use the categories of rated items and user interests in our training steps.
Then we can optimize the model by minimizing the following loss function: 

$\mathcal{L} = \mathcal{L}(w, w) + \mathcal{L}(u, w) + \mathcal{L}(p, w) + \mathcal{L}(p, u)$
where $w$, $u$, $p$ are the notations of words, users and rated items respectively.
Considering the large size of the vocabulary, users and rated items, we approximate our optimization objectives by the negative sampling.
Then we can treat each task as a classification problem and calculate loss values by the binary cross-entropy.
We present the details of each optimization task as following:

\paragraph{Word and word}
is a standard way to train Word2vec~\cite{mikolov2013distributed} models. The prediction task is to predict if the sampled words have co-occurred within the window context. The training process uses the negative sampling to approximate objective function. We choose 5 as the number of negative samples. We keep the top 20,000 frequent words and finally replace the rest with a special token, $<unk>$.

\paragraph{User and word}
predicts if a user authored the sampled words by the contexts of user posts. The goal is to learn patterns of user language usage from user historical posts. Given a document $i$, its author $u_i$ and the user's vocabulary $V_{u_i} = \{w_{1, i}, ..., w_{j, i}, ..., w_{n, i}\}$, where n is the number of frequent words authored by the user. Our objective is to minimize the following function:

$\mathcal{L}(u, w) = -\sum_{w_j \in V_{u_i} \\ w_k \in V \\ w_k \notin V_{u_i}}[log(\theta(e(u_i) \cdot e(w_j))) +  log(\theta(e(u_i) \cdot e(w_k)))]$
where $w_j$ is a negative sample from the whole vocabulary $V$, $e(u)$ and $e(w)$ are fixed-length user and word vectors respectively, and $\theta$ is a sigmoid function to normalize values of dot production. 
We extend the previous work~\cite{amir2017quantifying} to integrate both local and global user language usage by sampling $w_j$ from a combined token list of both the input document and the user's vocabulary.
This can help the model learn contextual information of each user.

\paragraph{Item and word}
follows the prediction task of user and word to classify if sampled words describe the selected item. This task is to use review documents to train representations of rated items. Then we can have 

$\mathcal{L}(p, w) = -\sum_{w_j \in V_{p_i}, w_k \in V, w_k \notin V_{p_i}}log(1 - \theta(e(p_i) \cdot e(w_j))) + log(\theta(e(p_i) \cdot e(w_k)))$
where $V_{p_i}$ is the vocabulary of the rated item $p_i$ and $w_k$ is a negative sample of words.
The language can be viewed as a bridge between an interactive relation of user and item, which predicts language usage for both rated items and users.

\paragraph{User and item}
learns if a user commented on the sampled items. The prediction task aims to adapt latent user factors into the user embeddings.
Given a document $i$, its author $u_i$ and the reviewed item $p_i$,
we can optimize the task by minimizing

$\mathcal{L}(u, p) = -\sum_{p_k \in P, p_k \notin P_{u_i}}log(\theta(e(p_i) \cdot e(u_i))) + log(\theta(e(p_k) \cdot e(u_i)))$
where the $P$ is a collection of all items, the $P_{u_i}$ is a reviewed item and the $p_k$ is a negative sample that the user does not review.
The constraints between reviewed items and users can help user embeddings identify language variations across domains of item genres.
And in turn, the relation of user and item can help infer item vectors.

For model settings, we used Adam~\cite{kingma2014adam} for the model optimization with a learning rate of 1e-5. We set the training epochs as 5. The model initializes embedding vectors randomly and learns 300-dimension representations for words, users and reviewed items. We empirically use 5 as the number of negative samples. For the other parameters, we keep the same as the defaults in the Keras~\cite{chollet2015keras}.

\section{Experiments}
\label{chap4:subsec:exp2}

We evaluate the effectiveness of the user factor adapted embedding model by an intrinsic evaluation, user clustering task and an extrinsic evaluation, personalized classification task. The first task aims to measure the purity of clusters with respect to categories of user interests, and the second task uses the document classification as a proxy of qualifying quality of user embeddings. We conduct a qualitative analysis of the user embeddings comparing with our close work~\cite{amir2017quantifying}.

\subsection{User Clustering Evaluation}

The unsupervised evaluation of embedding models focuses on four main categories: relatedness, analogy, categorization and selectional preference~\cite{schnabel2015evaluation}.
We approach the user embedding evaluation by categorizing users into different clusters. 
User communities or groups gather users by their interests and behaviors, such as engaging in the same filed of topics~\cite{benton2016learning, yang2017overcoming}.
In our datasets, the user-purchased Amazon products, the user-visited Yelp business units and the user-watched IMDb movies have their item categories. 
The categories can imply user preferences and interests, and therefore can help evaluate user clusters.
In this study, our proposed multitask model learns interactive relations across language, user and item instead of using the item categories. We compare our proposed model with other 5 baseline models:

\paragraph{word2user} 
represents users by aggregating word representations~\cite{benton2016learning}. We compute a user representation by averaging embeddings of all tokens that were authored by the user. To obtain the word embeddings, for each dataset, we trained a word2vec model for 5 epochs using Gensim~\cite{rehurek2010software} with 300-dimensional vectors.

\paragraph{lda2user} 
generates user representations by applying Latent Dirichlet Allocation (LDA)~\cite{blei2003latent} on user documents~\cite{pennacchiotti2011machine}. We set the number of topics as 300 and leave the rest of the parameters as their defaults in  Gensim~\cite{rehurek2010software}. We apply the LDA model on each user document to obtain a document vector, and then get a user vector by averaging the vectors of all the user's documents.

\paragraph{doc2user} 
applies paragraph2vec~\cite{le2014distributed} to obtain user vectors. We implemented the User-D-DBOW model which achieved the best performance in the previous work~\cite{ding2017multi}. The implementation keeps parameters with default values in the Gensim~\cite{rehurek2010software}. We aggregate each user's documents as a single document. Then the User-D-DBOW model can derive a single user vector from the aggregated document.

\paragraph{bert2user} 
follows a similar process of the lda2user. We use the ``bert-base-uncased'' pre-trained BERT model
for English from the transformers toolkit~\cite{Wolf2019HuggingFacesTS} with default parameter and model settings. After inserting ``[CLS]'' and ``[SEP]'' to the beginning and end of each document, the BERT model encodes a document into a fixed-length (768) document vector. We can then generate user embeddings by averaging each user's all document vectors.

\paragraph{user2vec} trains user embeddings by predicting word usage by users.
We follow the existing work~\cite{amir2017quantifying} but set the user vector dimension as 300.

\begin{table*}[htp]
\centering
\resizebox{\textwidth}{!}{
    \begin{tabular}{cc||ccc|ccc|ccc}
    \multicolumn{2}{c||}{\multirow{2}{*}{}} & \multicolumn{3}{c|}{Amazon-Health} & \multicolumn{3}{c|}{IMDb} & \multicolumn{3}{c}{Yelp} \\
    \multicolumn{2}{c||}{} & F1@4 & F1@8 & F1@12 & F1@4 & F1@8 & F1@12 & F1@4 & F1@8 & F1@12 \\\hline\hline
    \multirow{5}{*}{Baselines} & word2user & \textbf{.929} & .909 & .905 & .653 & .725 & .762 & .859 & .810 & .797 \\
     & lda2user & .920 & \textbf{.914} & .900 & .696 & .726 & .761 & .849 & .839 & .832 \\
     & doc2user & .873 & .891 & .901 & .660 & .725 & .748 & .836 & .828 & .826 \\
     & bert2user & .871 & .896 & \textbf{.906} & .660 & .714 & .734 & .838 & .828 & .830 \\
     & user2vec & .868 & .891 & .901 & .601 & .600 & .593 & .841 & .829 & .832 \\\hline
     
    Ours & MTL & .870 & .890 & .900 & \textbf{.801} & \textbf{.879} & \textbf{.884} & \textbf{.879} & \textbf{.843} & \textbf{.838} 
    \end{tabular}
}
\caption{Performance summary of different user embedding models. We report F1 scores at multiple numbers of clusters. The bold fonts indicate the best performance in each evaluation task.}
\label{chap4:tab:usereval}
\end{table*}

We use the \texttt{SpectralClustering} algorithm from scikit-learn~\cite{pedregosa2011scikit} toolkit to cluster users into three clustering sizes, 4, 8 and 12. 
We set the affinity as cosine and leave other parameters as their defaults.
To measure cluster quality, we select every two users from the clusters without repetition.
We count the user pair as a correct option if two users have overlaps within the same item genre and from the same cluster or if the user pair does not overlap and is from the different clusters.
Otherwise, we will count the selection as the wrong option.
Therefore, we can have a list of predicted labels and ground truths by using the item genres as a proxy.
Finally, we measure the clustering purity by the F1 score.

We present results at Table~\ref{chap4:tab:usereval}.
The results show that our multitask user embedding model outperforms the other baselines by a large portion on the IMDb and Yelp datasets.
The improvements suggest the user factor adapted model can understand semantic variations in diverse user interests.
The performance of our model and user2vec has similar scores on the Amazon-Health dataset.
Comparing to the other two datasets, the Amazon-Health data has more similar topics of review items.

\subsection{Personalized Classifier Evaluation}

We train three classifiers to evaluate user embeddings on the document classification task.
We split each dataset into training (80\%), development (10\%) and test (10\%) sets, as shown in Table~\ref{chap4:tab:data2}. 
The models oversample the minority during the training process.
We test the classifiers when they achieve the best performance on the development set.
Finally, we report precision, recall and F1 scores using the \texttt{classification\_report} from scikit-learn~\cite{pedregosa2011scikit}.
Figure~\ref{chap4:fig:uemb_diagram} illustrates personalizing classifiers by concatenating document representations with user embeddings.
We compare our proposed model with classifiers using existing user2vec~\cite{amir2017quantifying} and non-personalized classifiers.
To ensure a fair comparison, classifiers use the same settings for models with and without user embeddings.

\paragraph{LR.}
We build a logistic regression classifier using \texttt{LogisticRegression} from scikit-learn \cite{pedregosa2011scikit}. The classifier extracts uni-, bi- and tri-gram features on the corpora with the most frequent 15K features with default parameters.

\paragraph{GRU.} 
We build a bi-directional Gated Recurrent Unit (GRU)~\cite{cho2014properties} classifier. We padded documents to the average document length of each corpus. We set the output dimension of GRU as 200 and apply a dense layer on the output. The dense layer uses ReLU~\cite{hahnloser2000digital} as the activation function, applies a dropout~\cite{srivastava2014dropout} rate of 0.2 and outputs 200 dimensions for final document class prediction. We train the classifier for 20 epochs.

\paragraph{BERT.} 
We implement a BERT-based classifier by HuggingFace's transformers toolkit \cite{Wolf2019HuggingFacesTS}. The classifier loads the ``bert-base-uncased'' pre-trained BERT model for English, encodes each document into a fixed-length (768) vector and feeds to a linear prediction layer for prediction.  We conduct fine-tuning steps for 10 epochs with a batch size of 32 and optimize the model by \texttt{AdamW} with a learning rate of 9$e^{-5}$.

\begin{table*}[t]
\centering
\begin{tabular}{c||ccc|ccc|ccc}
\multirow{2}{*}{Methods} & \multicolumn{3}{c|}{Amazon-Health} & \multicolumn{3}{c|}{IMDb} & \multicolumn{3}{c}{Yelp} \\
 & Precision & Recall & F1 & Precision & Recall & F1 & Precision & Recall & F1 \\\hline\hline
LR & .834 & .768 & .793 & .818 & .779 & .794 & .856 & .820 & .833 \\
LR-u & .841 & .777 & \textbf{.801} & .834 & .791 & \textbf{.807} & .860 & .821 & .835 \\
LR-up & .838 & .771 & .796 & .833 & .791 & \textbf{.807} & .863 & .825 & \textbf{.838} \\\hline
GRU & .813 & .844 & .812 & .824 & .837 & .823 & .851 & .865 & .852 \\
GRU-u & .836 & .811 & .821 & .832 & .819 & .825 & .868 & .846 & .858 \\
GRU-up & .821 & .832 & \textbf{.825} & .846 & .824 & \textbf{.836} & .876 & .864 & \textbf{.867} \\\hline
BERT & .866 & .822 & .840 & .852 & .809 & .826 & .866 & .825 & .840 \\
BERT-u & .863 & .812 & .831 & .858 & .818 & .833 & .872 & .843 & \textbf{.854} \\
BERT-up & .873 & .838 & \textbf{.851} & .864 & .831 & \textbf{.844} & .880 & .839 & \textbf{.854}
\end{tabular}
\caption{Performance scores of document classifiers on the review datasets. `-u` means personalized classifiers using user2vec~\cite{amir2017quantifying} and `-up` indicates personalizing classifiers via our proposed method. We use the bold fonts to highlight the best performance of each classifier on separate datasets.}
\label{chap4:tab:personalization}
\end{table*}

We show the performance results in Table~\ref{chap4:tab:personalization}. Comparing to the baselines, the classifiers personalized by our proposed model generally achieve the best performance across the three datasets. 
This highlights adapting user factors can help embedding models learn user variations and benefit the classification performance.
We can also observe that the personalized classifiers generally outperform the non-personalized classifiers.
This indicates personalizing the classifiers with user history boosts classification performance in our study.

\subsection{Visualization Analysis}

\begin{figure*}[t]
\centering
\includegraphics[width=0.44\textwidth]{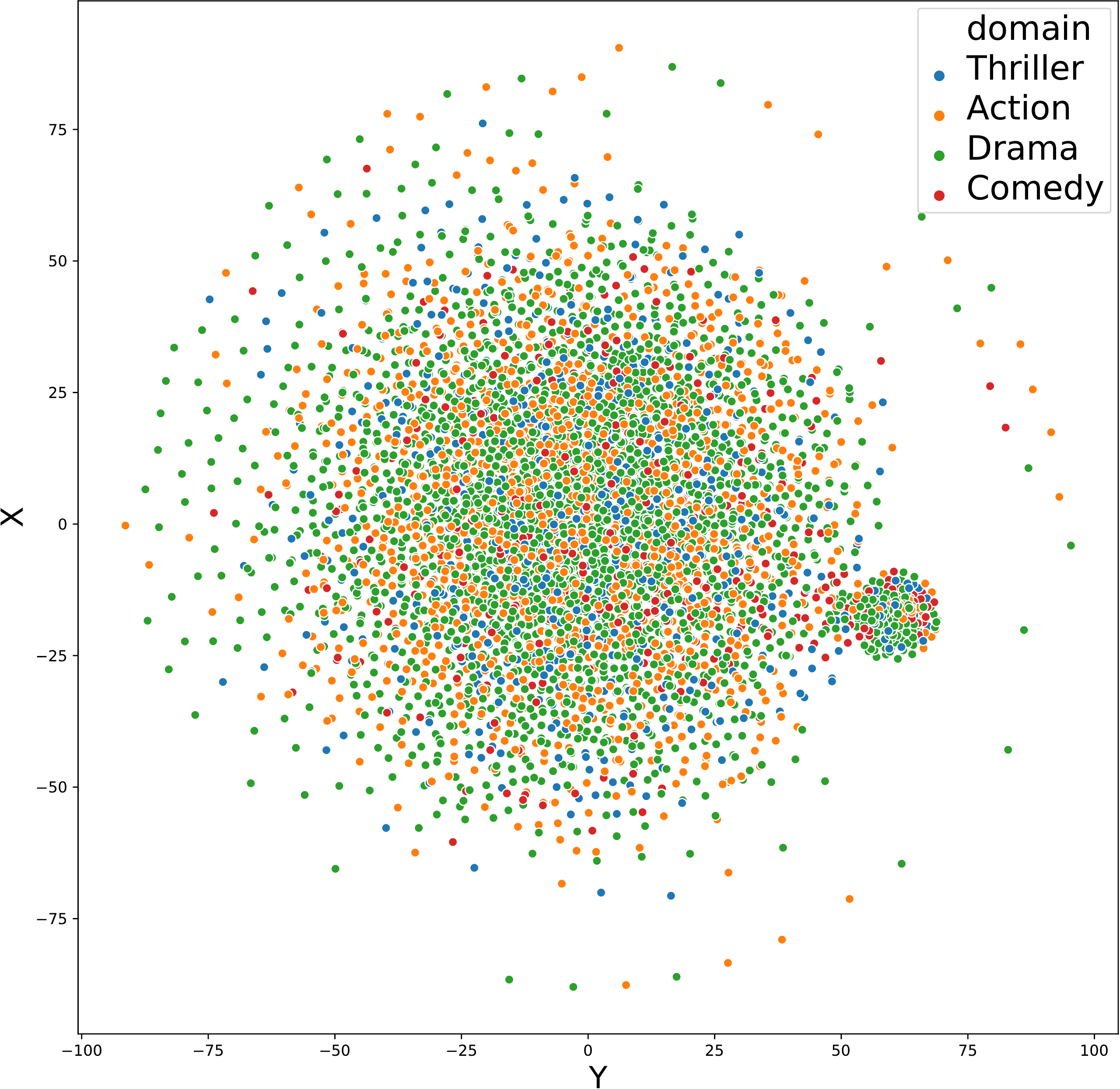}
\includegraphics[width=0.44\textwidth]{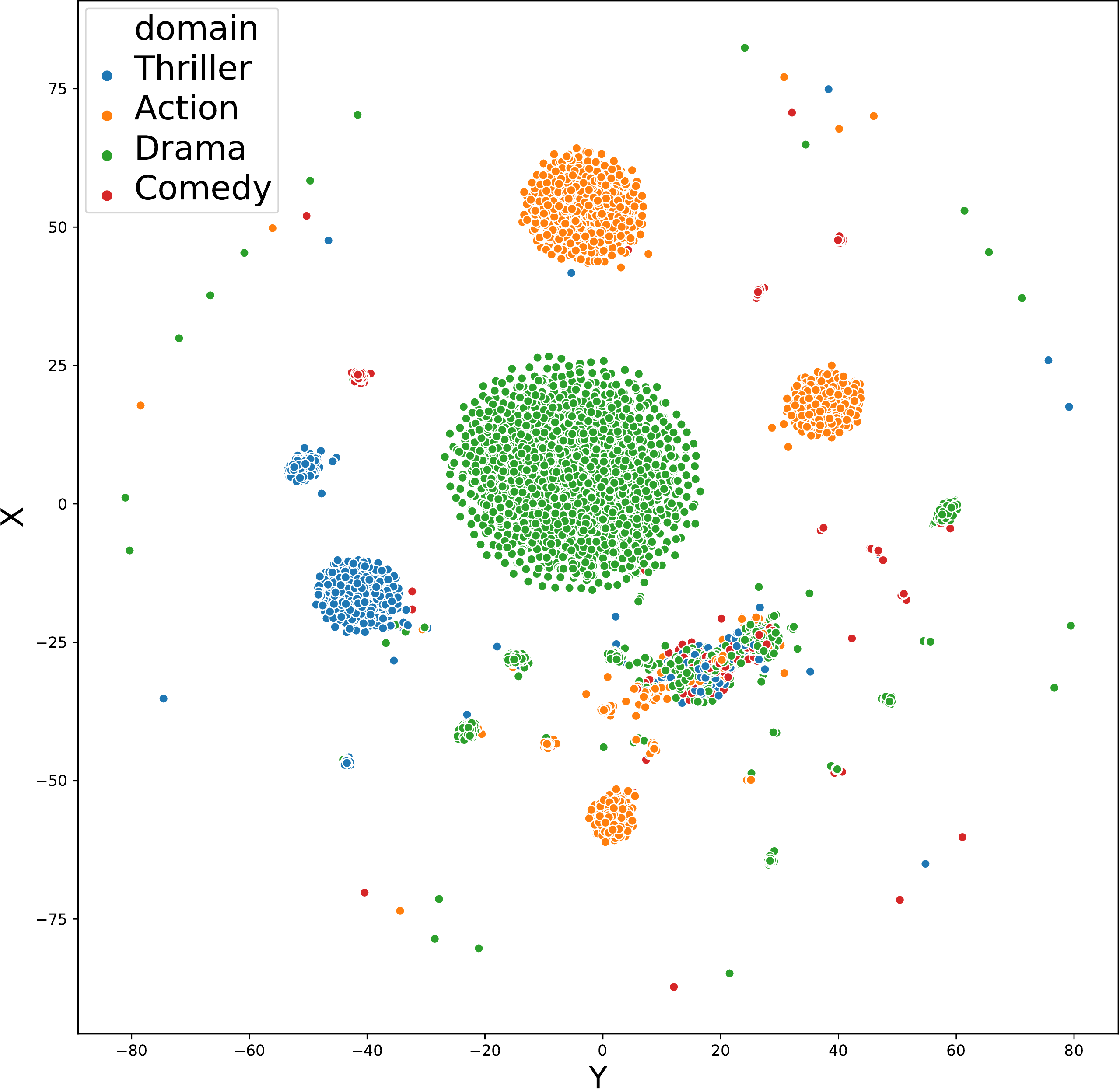}
\caption{Visualizations of IMDb users colored concerning their interests in 4 movie genres. We plot users using the embeddings from our proposed method (right) and user2vec~\cite{amir2017quantifying} (left). The visualizations of Yelp and Amazon are omitted for reasons of space.}
\label{chap4:fig:uemb_viz}
\end{figure*}

To further evaluate the effectiveness of user embedding models, we map users into a 2-D space using user embeddings and plot them in Figure~\ref{chap4:fig:uemb_viz}.
We group users according to user interests using the domain categories of rated items.
To map the 300-d user embeddings, we use the \texttt{TSNE} algorithm from scikit-learn~\cite{pedregosa2011scikit} to compress the dimension into 2-d vectors. We set the n\_component as 2 and leave the other parameters as their defaults in the TSNE.
We can observe that the MTL user embedding model shows more clustering patterns with regard to user interests (categories of reviewed items).
This indicates that the unsupervised multitask learning framework can adapt the latent user factors into the user embedding.
Users may have multiple interests. In the right plot, we can also find that there is a cluster that mixes with multiple colors on the right bottom.

\section{Related Work}

\paragraph{User Profiling} is a common task in natural language processing.
Online generated user texts show demographic variations in the linguistic styles, and the linguistic style variability could be used for predicting user's personality and demographic attributes~\cite{rosenthal2011age, zhang2016predicting, hovy2018improving, wood2020using, gjurkovic2020pandora, lynn2020hierarchical}. 
The demographic user factors influence how online users express their opinions~\cite{volkova2013exploring, hovy2015demographic, wood2017does} and show promising improvements in the text classification task~\cite{lynn2017human, huang2019neuraluser, lynn2019tweet}.
However, in this work, the goal of modeling user factor is to train robust user embeddings via domain adaptation, rather than the end goal being demographic factor prediction and document classification itself.

\paragraph{Personalized classification} generally improves the performance of document classifiers~\cite{flek2020returning}.
The multitask learning framework has been applied for personalizing document classifiers by optimizing the classifiers on multiple document levels~\cite{benton2017multitask} or general and individual levels~\cite{wu2016personalized}.
The social relation can bridge connections between users and generalize classification models across users~\cite{wu2016personalized, yang2017overcoming}. 
For example, \cite{wu2016personalized} optimizes document classifiers by two optimization tasks, sentiment classification and user social relation minimization, which allows classifiers to minimize the impacts of user community variations.
This work personalizes classifiers in a different way, where we train user embedding models under a multitask learning framework and use the personalized classifiers to evaluate user embedding models.

\section{Conclusion}

In this study, we have proposed user factor adaptation for building user embedding under a multitask framework.
Our analyses show how the user factor causes semantic variations in relation to word usage and document classification, showing that the user factor is rooted in language.
We have evaluated the proposed user embedding models in both intrinsic and extrinsic tasks. 
The user factor adapted model has shown its robustness to language variations in both instrinsic and extrinsic evaluations, learning user representations and personalizing classifiers.
We release our source code and instructions of data access at \url{https://github.com/xiaoleihuang/UserEmbedding}.

Our work in user factor adaptation highlights several future directions to explore.
First, our method models latent user factors inferred from user posts. A combination of user embedding and explicit attributes (e.g., demographic factors) may improve model personalization.
Second, user behaviors shift over time.
A time-adapted user embedding can jointly model temporality and user attributes in online social media and can be extended to other fields, such as public health.

\section{Acknowledgement}
The authors thank the anonymous reviews. This work was supported in part by the National Science Foundation under award number IIS-1657338. This work was also supported in part by a research gift from Adobe Research. The first author would thank the computational support from the JHU CLSP cluster.

\bibliography{eacl2021}
\bibliographystyle{acl_natbib}

\end{document}